\documentclass[10pt,twocolumn,letterpaper]{article}

\usepackage[pagenumbers]{cvpr} 

\usepackage[dvipsnames]{xcolor}

\definecolor{cvprblue}{rgb}{0.21,0.49,0.74}
\usepackage[pagebackref,breaklinks,colorlinks,citecolor=cvprblue]{hyperref}

\def\paperID{*****} 
\def\confName{CVPR}
\def\confYear{2024}

\title{1\textsuperscript{st} Place Winner of the 2024 Pixel-level Video Understanding in the Wild (CVPR'24 PVUW) Challenge in Video Panoptic Segmentation \\ and Best Long Video Consistency of Video Semantic Segmentation.}   

\author{
Qingfeng Liu\thanks{ Corresponding author.}{\qquad}Mostafa El-Khamy{\qquad}Kee-Bong Song\\
{Device Solutions Research America}, Samsung Semiconductor, Inc., San Diego, CA 92121 \\
qf.liu@samsung.com, m\_elkhamy@ieee.org, keebong.s@samsung.com 
}

\begin{document}
\maketitle
\begin{abstract}
 The third Pixel-level Video Understanding in the Wild (PVUW CVPR 2024) challenge aims to advance the state of art in video understanding through benchmarking Video Panoptic Segmentation (VPS) and Video Semantic Segmentation (VSS) on challenging videos and scenes introduced in  the large-scale Video Panoptic Segmentation in the Wild (VIPSeg) test set and the large-scale Video Scene Parsing in the Wild (VSPW) test set, respectively. This paper details our research work that achieved the 1\textsuperscript{st} place winner in the PVUW'24 VPS challenge, establishing state of art results in all metrics, including the Video Panoptic Quality (VPQ) and Segmentation and Tracking Quality (STQ). With minor fine-tuning our approach also achieved the 3\textsuperscript{rd} place in the PVUW'24 VSS challenge ranked by the mIoU (mean intersection over union) metric and the first place ranked by the VC16 (16-frame video consistency) metric. Our winning solution stands on the shoulders of giant foundational vision transformer model (DINOv2 ViT-g) and proven multi-stage Decoupled Video Instance Segmentation (DVIS) frameworks for video understanding. 
\end{abstract}    
\section{Introduction}
\label{sec:intro}

\begin{figure*}[t]
  \centering
   \includegraphics[width=1\linewidth]{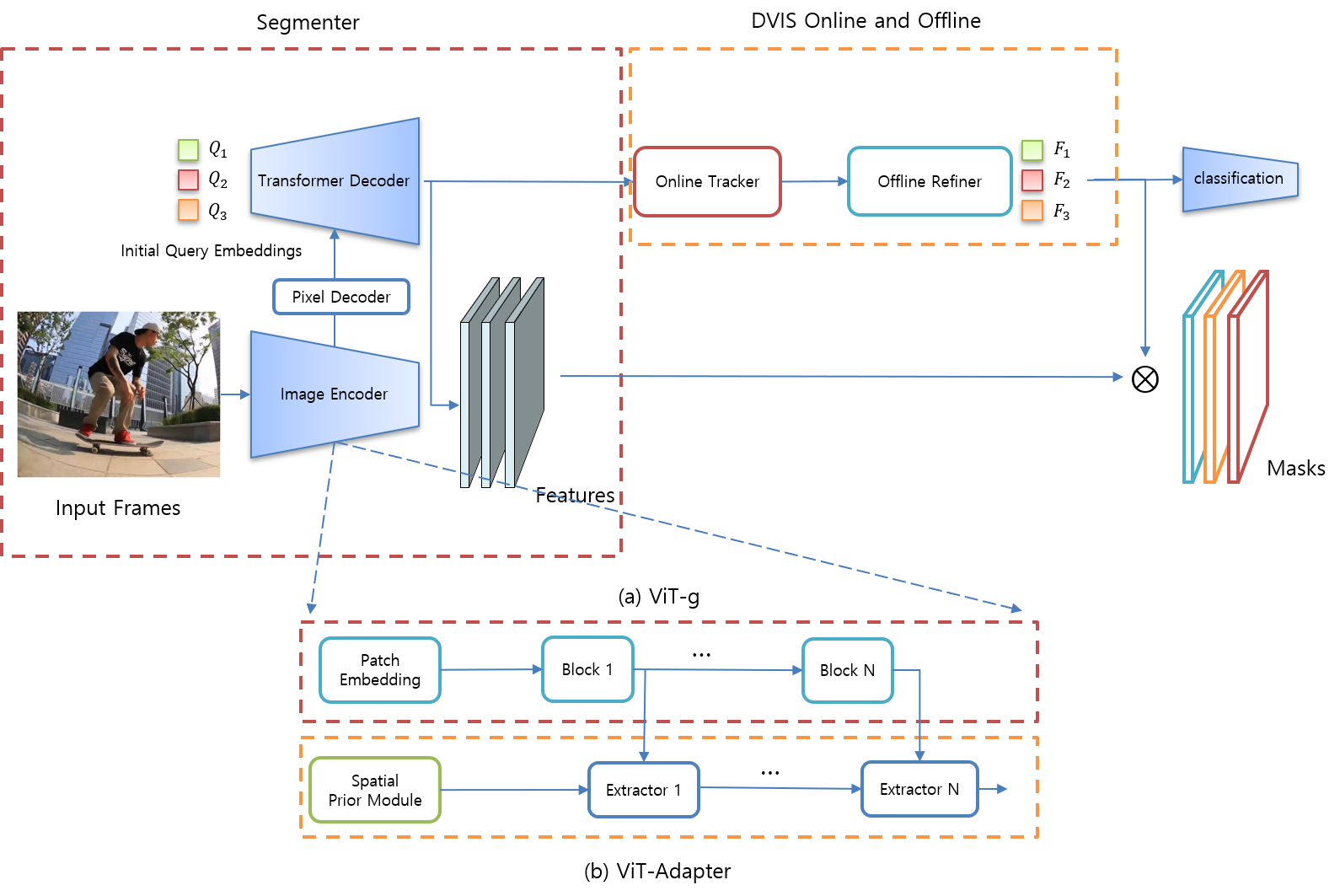}
   \caption{\textbf{The System architecture overview.} It consists of three independent components: a segmenter, an online tracker, and an offline temporal refiner. We mainly focus on improve the performance of segmenter by applying ViT-Adapter with ViT-g pretrained by DINOv2.
   }
   \label{fig:framework}
\end{figure*}


Video Panoptic Segmentation (VPS)~\cite{vps} is an important computer vision task for scene understanding that is likely to be widely deployed for autonomous driving, virtual reality, and mobile phone intelligent cameras. 
It aims at identifying, segmenting, classify, and simultaneously tracking all instances for classes of interest and background stuff classes across all frames in a video. 

Whereas image panoptic segmentation segments and classifies instances of interest and background stuff classes, VPS is more complex since it must provide consistent segmentation and accurate tracking of  each segmented instance across all consecutive video frames it appears in. 
This becomes particularly challenging with complex background scenes, multiple moving subjects, occlusions, changed camera angles, and longer videos. 
In particular for videos with long duration, the same instance or background class may exhibit significant variations in positions, poses, shapes, occlusions, and view angles, which complicates the challenge of accurately associating the same ID across frames that are far apart. 

Earlier work of VPS extended the models of image panoptic segmentation \cite{panoptic-deeplab, knet, kmaxdeeplab} to video domain \cite{vps, vip-deeplab, video-knet, tubelink, videokmax}.
Recent work focuses on developing universal models that be applied to all video segmentation tasks, including Video Semantic Segmentation (VSS) and Video Instance Segmentation (VIS).
State-of-the-art (SOTA) VPS methods ~\cite{minvis, dvis, dvisplusplus, maxtron} have demonstrated promising performance by refining the embeddings generated from SOTA image segmentation models~\cite{mask2former,kmaxdeeplab} using the temporal information.

One important work is DVIS \cite{dvis} that decoupled the task of video panoptic segmentation into three independent sub-tasks: image segmentation, online tracker, and offline refiner. DVIS++~\cite{dvisplusplus} further proposed a Noiser module to enhance the online tracker to acquire more stable and powerful tracking capabilities.
Recent trend is to investigate the performance of large visual foundation models on the down-streaming tasks. DVIS++~\cite{dvisplusplus} used a medium size foundation model DINOv2-L~\cite{dinov2} (the large version of ViT trained by DINOv2) that improves the performance upon Swin-L~\cite{swin} significantly. InternVL~\cite{chen2023internvl} has shown impressive results on semantic segmentation using its 6b parameter model InternViT-6B.

Inspired by state-of-the-art methods, we conducted extensive experiments and proposed to integrate the ViT-Adapter with DINOv2-g (the giant version ViT trained by DINOv2) into the DVIS framework. As a result, using this 1.2 billion parameter model, we achieved the first place in the VPS track of the PVUW challenge in CVPR 2024. Specifically, our model achieved 58.26\% VPQ at the final test phase using single scale inference without any test time augmentation and model ensemble, as well as without using any additional training data (including the validation set of VIPSeg).
In addition, we also achieved  mIoU 63.92\%, VC8 94.84\% and VC16 93.25\% in the test phase of VSS track, and ranked third place by finetuning our model trained on VPS track on VSS dataset.
\section{Method}
DVIS proposed a novel decoupled framework for Video Panoptic Segmentation that consists of three independent components: an image segmenter, an online tracker, and an offline temporal refiner, as illustrated in \ref{fig:framework}. For more details, one can refer the DVIS paper~\cite{dvis}.
DVIS employed Mask2Former \cite{mask2former} as the segmenter in our framework. Mask2Former is constructed upon a straightforward meta-architecture comprising a backbone, a pixel decoder, and a transformer decoder.

We found that DVIS already showed impressive results in terms of refining the embeddings, and marginal gain or no gain are observed by increasing the number of blocks in the online tracker and offline temporal refiner in our experiments.
DVIS++ used a medium size foundation model DINOv2-L (the large version of ViT trained by DINOv2) and a reduced ViT-Adapter as the backbone that improves the performance upon Swin-L significantly.

Therefore, we further scale up the size of backbone by applying the ViT-Adapter using DINOv2-g inspired by DVIS++, for performance improvement.
Please note that we neither used the Noiser module in DVIS++ as the training performance is not stable in our case, nor used the contrastive loss used in DVIS++ since it consumes a lot of GPU memory.

\subsection{Large Visual Foundation Models \label{sec:lvfm}}

The backbone network is crucial to the performance for Video Panoptic Segmentation.
we started with different backbones and investigated their performance on the VIPSeg validation dataset.
In particular, we showed performance on VIPSeg validation dataset using various backbones as shown in Table~\ref{tab:vipseg} including lightweight backbone MobileViTv2~\cite{mobilevitv2}, medium size backbone Swin-L~\cite{swin}, DINOv2-L~\cite{dinov2} and large visual foundation backbone DINOv2-g~\cite{dinov2}. 
we also considered a even larger size backbone such as the InternViT-6B. However, for the challenge we limited our submissions to results using the DINOv2 ViT-g foundational model as the backbone. 
We can observe that With large visual foundation backbone DINOv2-g, we can achieve the state-of-the-art performance 60.42\% VPQ on VIPSeg validation dataset.

\begin{table}[t]
\centering
 \caption{\textbf{Results on the VIPSeg validation set.$^*$ denotes our training} } 
 \label{tab:vipseg}
\begin{tabular}{l | l | c }
	Method & Backbone & VPQ \%  \\
	\hline
	Video K-Net~\cite{video-knet}& ResNet-50 & 26.1  \\
	TarVIS~\cite{tarvis}& ResNet-50 & 33.5  \\
         Tube-Link~\cite{tubelink}& ResNet-50 & 39.2 \\
         Video-kMax\cite{videokmax}& ResNet-50 & 38.2  \\
         DVIS~\cite{dvis} & ResNet-50 & 43.2  \\
         DVIS++~\cite{dvisplusplus} & ResNet-50 & 44.2  \\
	\hline
	TarVIS~\cite{tarvis} & Swin-L & 48.0 \\
	DVIS~\cite{dvis} & Swin-L & 57.6 \\ 
	DVIS++~\cite{dvisplusplus} & DINOv2-L & 58.0 \\
	\hline
    DVIS$^*$ & MobileViTv2 & 36.94 \\
    DVIS$^*$ & Swin-L & 57.81 \\
    DVIS$^*$ & DINOv2-L & 58.80 \\
    DVIS$^*$ & DINOv2-g & \textbf{60.42} \\
\hline
 \end{tabular}
\end{table}

\subsection{ViT-Adapter design\label{sec:vit_adapter}}

Current large visual foundation models are normally plain ViT, that is pretrained on massive scale dataset to learn semantic-rich representations.
However, unlike specifically designed backbone like Swin, the plain ViT normally has inferior performance on dense prediction tasks due to the lack of multiple scale features that are very important to performance.

We reference the design of the ViT-Adapter~\cite{vitadapter} used in DVIS++~
\cite{dvisplusplus}, as demonstrated in Figure~\ref{fig:framework}. 
We adapt the embedding from the large foundation model through the ViT-Adapter, that is implemented as an additional branch that interacts with the plain ViT blocks with a spatial prior module and several Extractor modules.
Note that original ViT-Adapter~\cite{vitadapter} also has several Injector modules, which we removed, since, in our approach, the large visual foundational backbone ViT-g is frozen during training.

The input image is input into the Spatial Prior module, where multiple scales of feature maps (e.g. 1/4, 1/8, 1/16, and 1/32) will be extracted and concatenated as input for Extractors.
The role of the ViT-Adapter becomes evident, as we then obtain multiple-scale features by splitting the ViT-Adapter output and providing them as the input for the pixel decoder. The ViT-Adapter is trained together with the pixel decoder and transformer decoder in an end to end fashion, while freezing the DINOv2 ViT-g (DINOv2-g) backbone. 

In our experiments, the DINOv2-g consists of 40 blocks, we divide them evenly into four stages by indices [[0, 9], [10, 19], [20, 29], [30, 39]], where the output of each stage will interact with the ViT-Adapter by using multi-scale deformable attention, where we set the number of heads as 24.

We also considered another design choice that we call ViT-CoMer-Adapter, which is a GPU memory friendly combined version of ViT-CoMer~\cite{vitcomer} and ViT-Adapter, where the Multi-Receptive Field Feature Pyramid (MRFP) module~\cite{vitcomer} is applied before each Extractor module for better multiple scale feature interactions.
However, in our preliminary experiments, ViT-CoMer-Adapter achieved similar performance as ViT-Adapter shown in Table~\ref{tab:coco}

\begin{table}[t]
\centering
 \caption{\textbf{Results on the COCO Panoptic Segmentation validation set} } 
 \label{tab:coco}
\begin{tabular}{l | c }
\hline
	Method & PQ \%  \\
 \hline
    ViT-Adapter & 59.10 \\
    ViT-CoMer-Adapter & 59.01 \\
\hline
 \end{tabular}
\end{table}

\subsection{Finetune for Video Semantic Segmentation}

To test the generalization capability, we tested the model pretrained on VIPSeg dataset for the VPS task directly on the VSPW dataset for the VSS task without any finetuning. For this, we simply merged all the predicted instances from the same class into a single class segmentation mask for the VSS task.
Next, we also finetuned the model pretrained on VIPSeg dataset on VSPW dataset.
It is interesting to observe that the pretrained model achieved better temporal consistency VC8 and VC16 on VSPW test dataset without any finetuning, while the finetuned model achieved better mIoU, as shown in Table~\ref{tab:vss_finetune}. This can be attributed to the fact that video panoptic segmentation puts more emphasis on video consistency by the extensive training for accurately tracking multiple instances across frames. However, video semantic segmentation training puts more emphasis on the boundaries of each class, hence improves the per-frame mean intersection over union (mIoU) metric.

\begin{table}[t]
\setlength{\tabcolsep}{1.4mm}
\centering
\begin{tabular}{l|cccc}
	Method & mIoU & weighted IoU & VC8 & VC16   \\
	\hline
    no finetuning & 0.6290 & 0.7431 & \textbf{0.9498} & \textbf{0.9332}  \\
    fintuned & \textbf{0.6392} & \textbf{0.7447} & 0.9484 & 0.9325 \\
\hline
 \end{tabular}
\caption{\textbf{Comparison with fintuned model and pretrained model without finetuning for VSS track on the test phase dataset.}}
 \label{tab:vss_finetune}
\end{table}

\section{Experiment}
\begin{table*}[t]
\setlength{\tabcolsep}{1.4mm}
\centering
\begin{tabular}{l|cccccc}
	Method & VPQ & VPQ1 & VPQ2 & VPQ4 & VPQ6 & STQ   \\
	\hline
    SiegeLion & 56.36 & 57.14 & 56.46 & 56.03 & 55.80 & 0.5252 \\
    kevin1234 (ours) & 55.69 & 56.41 & 55.86 & 55.39 & 55.11 & 0.519 \\
    Reynard & 54.55 & 55.27 & 54.69 & 54.25 & 53.97 & 0.517 \\
    ipadvideo & 54.26 & 54.96 & 54.44 & 53.98 & 53.65 & 0.509 \\ 
    zhangtao-whu & 52.77 &	53.32 & 52.92 & 52.57 & 52.26 & 0.502 \\
\hline
 \end{tabular}
\caption{\textbf{Leaderboard of VPS track during the development phase.}}
 \label{tab:develop}
\end{table*}

\begin{table*}[t]
\setlength{\tabcolsep}{1.4mm}
\centering
\begin{tabular}{l|cccccc}
	Method & VPQ & VPQ1 & VPQ2 & VPQ4 & VPQ6 & STQ   \\
	\hline
	\textbf{kevin1234 (ours)} & \textbf{58.26} & \textbf{59.10} & \textbf{58.50} & \textbf{57.90} & \textbf{57.53} & \textbf{0.5434} \\
    SiegeLion & 57.12 & 58.21 & 57.41 & 56.68 & 56.17 & 0.5397 \\
    Reynard & 57.01 & 57.89 & 57.22 & 56.65 & 56.28 & 0.5343 \\
    ipadvideo & 28.38 & 29.12 & 28.68 & 28.10 & 27.63 & 0.2630 \\
    JMCarrot & 22.11 & 23.81 & 22.83 & 21.49 & 20.29 & 0.2603 \\
\hline
 \end{tabular}
\caption{\textbf{Leaderboard of VPS track during the test phase.}}
 \label{tab:test}
\end{table*}

\subsection{Dataset}

The VIPseg and VSPW datasets are used for the Video Panoptic Segmentation track and Video Semantic Segmentation track in PVUW challenge 2024.

\textbf{VIPSeg} VIPSeg~\cite{vipseg} dataset is a large-scale dataset for Video Panoptic Segmentation dataset in the wild. There are 124 classes including 58 things’ and 66 stuff’s classes. It consists of 3,536 videos and 84,750 frames, where 2806/343/387 videos are for training/validation/testing, respectively.

\textbf{VSPW} VSPW~\cite{miao2021vspw} dataset has the same set of images for training/validation/testing as VIPSeg dataset, while the annotation is just for video semantic segmentation.

\subsection{Implementation Details}

For VPS, we follow the same training pipeline as the Decoupled Video Instance Segmentation (DVIS) frameowrk. The segmenter, referring tracker, and temporal refiner are trained separately.
We first pretrained the Mask2former model~\cite{mask2former} as our segmenter on MSCOCO panoptic segmentation dataset~\cite{coco} using batch size 16 and learning rate 1e-4 for 50 epochs. 
Since we freeze the weights of the large visual foundation model DINOv2-g, as the backbone for feature embeddding, we only train the ViT-adapter part and the transformer decoder part in this stage.
Then, we finetuned the segmenter using image-level annotations from the training set of VIPSeg, following the training strategy of MinVIS~\cite{minvis} with batch size 16 and learning rate 1e-4 for 20000 iterations, with a learning-rate decay rate of 0.1 at 14000 iteration.
With the initialization of segmenter, we further trained the DVIS online tracker with batch size 8 and learning rate 1e-4 for 40000 iterations while reducing the learning rate by half at 28000 iteration. In the last training stage, the DVIS offline temporal refiner is trained using the same setting on VIPSeg training datasets.

We also explored to increase the crop size of the DVIS offline training from (604, 604) to (840, 840), so we finetuned our model using batch size 8 and learning rate 2.5e-5 for 10000 iterations and decayed at 7000 iteration using the model trained with crop size (604, 604) as initialization.

All the trainings are conducted on single machine of 8x A100 Nvidia GPUs. 
The DVIS online model used a consecutive 5-frame clip from the video as input. The DVIS offline model used a window size of 21 frames as input.
Multi-scale training is used to randomly scale the short side of input video clips from 480 to 1080 during training. 
By default, a random crop size (604, 604) is used except the last stage of finetuning using crop size (840, 840).

For VSS, we used the best model pretrained on VIPSeg dataset as initialization and finetuned it on the VSPW dataset using a batch size of 8 and learning rate 2.5e-5 for 10000 iterations and decayed at 7000 iteration.

\subsection{Results for Video Panoptic Segmentation Track}

In the VPS track of PVUW Challenge 2024, we ranked first in the test phase. The leaderboards for the development and test phases are displayed in Table~\ref{tab:develop} and Table~\ref{tab:test}, respectively. Our method achieved a VPQ of 55.69 in the development phase and 58.26 in the test phase. 

With the help of large visual foundation model as the backbone, our method surpassed all other methods in all metrics in the test phase.
Additionally, our method also showed its tracking stability when comparing the performance drop between VPQ6 and VPQ1.
One thing to note that we only used single scale inference to achieve the result without using any test time augmentation or model ensemble.

\subsection{Results for Video Semantic Segmentation Track}

In the VSS track of PVUW Challenge 2024, we ranked 3rd in the test phase, ranked by the mIoU metric and first ranked by the VC16 (16-frame video-consistency) metric. The leaderboard for the test phase is displayed in Table~\ref{tab:vss_test}. Our method achieved a mIoU of 55.69 i and VC16 of 0.9325, in the PVUW VSS test phase. 

\begin{table}[t]
\setlength{\tabcolsep}{1.4mm}
\centering
\begin{tabular}{l|cccc}
	Method & mIoU & weighted IoU & VC8 & VC16   \\
	\hline
    SiegeLion & 0.6783 & 0.7761 & 0.9482 & 0.9290  \\
    lieflat & 0.6727 & 0.7659 & 0.9499 & 0.9312  \\
    \textbf{kevin1234 (ours)} & 0.6392 & 0.7447 & 0.9484 & \textbf{0.9325} \\
    bai\_kai\_shui & 0.6375 & 0.7422 & 0.9462 & 0.9290  \\
    JMCarrot & 0.6337 & 0.7456 & 0.9462 & 0.9294  \\
    ipadvideo & 0.5854 & 0.7145 & 0.9073 & 0.8802  \\
\hline
 \end{tabular}
\caption{\textbf{Leaderboard of VSS track during the test phase.}}
 \label{tab:vss_test}
\end{table}

In particular, our model tends to have better temporal consistency in terms of metrics VC8 and VC16. We ranked 2nd in VC8 and ranked 1st in VC16, which means our model tends to have better tracking reliability since it is finetuned from a VIPSeg pretrained model.
We believe that better mIoU can be achieved if we follow the same training pipeline as VIPSeg dataset, rather than simply finetuning a VIPSeg pretrained model on VSPW dataset.

\subsection{Ablation Studies}

We also carefully tuned the hyper-parameters for several aspects using the development stage dataset.
As shown in Table~\ref{tab:crop_size}, a finetuned model with crop size (840, 840) achieved slightly better performance than baseline (604, 604) and also (630, 1120) with a different aspect ratio 16:9.
We also compared different query sizes and found that 300 performs similar as 200, so we still keep 200 query size as our choice, as shown in Table~\ref{tab:query_size}.

\begin{table}[t]
\setlength{\tabcolsep}{1.4mm}
\centering
\begin{tabular}{l|cccccc}
	Crop size & VPQ & VPQ1 & VPQ2 & VPQ4 & VPQ6 & STQ\\
    \hline
    (604, 604) & 55.18 & 55.99 & 55.29 & 54.79 & 54.67 & 0.515\\
    (630, 1120) & 55.62 & 56.30 & 55.77 & 55.34 & 55.08 & 0.519 \\
	(840, 840) & 55.69 & 56.41 & 55.86 & 55.39 & 55.11 & 0.519\\
\hline
 \end{tabular}
\caption{\textbf{Comparison for different crop sizes.}}
 \label{tab:crop_size} \vspace{-2mm}
\end{table}

\begin{table}[t]
\setlength{\tabcolsep}{1.4mm}
\centering
\begin{tabular}{l|cccccc}
	Query size & VPQ & VPQ1 & VPQ2 & VPQ4 & VPQ6 & STQ\\
    \hline
    300 & 55.27 & 55.95 & 55.41 & 54.97 & 54.76 & 0.516 \\
	200 & 55.69 & 56.41 & 55.86 & 55.39 & 55.11 & 0.519\\
\hline
 \end{tabular}
\caption{\textbf{Comparison for different query sizes.}}
 \label{tab:query_size} \vspace{-2mm}
\end{table}

\section{Discussions and Conclusions}
This work demonstrates how the embeddings from large foundational vision models such as the 1.1B-parameter DINOv2 ViT-g model can be adapted for very complex vision understanding tasks such as Video Panoptic Segmentation (VPS) and Video Semantic Segmentation (VSS). By adapting the foundational vision model embeddings to the multistage decoupled video instance segmentation (DVIS) framework, our solution achieved the 1\textsuperscript{st} place winner in the CVPR'24 PVUW VPS challenge, establishing new state of the art 58.26 VPQ (Video Panoptic Quality) and 0.543 STQ (Segmentation and Tracking Quality) on the VIPSeg test set. For the CVPR'24 PVUW VSS challenge, by further fine-tuning, our same solution achieved  3\textsuperscript{rd} place in the mIoU metric with 0.64 mIoU  and 1\textsuperscript{st} place in 16-frame video consistency metric with 0.933 VC16, on the challenging VSS test set.

The crux of our solution is carefully designing and training a vision transformer adapter (ViT-Adpater) that adapts the embedding from large foundational models for complex downstream vision tasks, while freezing the large foundational models. We showed by only training the ViT-adapter with the multistage down-stream vision models, this method generalizes well to multiple tasks. We demonstrated that the same embeddings from the frozen giant ViT-g can be adapted for either the Video Panoptic Segmentation task or the Video Semantic Segmentation task, while achieving the state of art results in both tasks. This approach of sharing a foundational backbone across multiple vision tasks, has important implications in the design of deep-learning based vision systems,  such as the reduction in training costs, simplified deep-learning vision model design,  as well as reduced inference computational costs and inference latency for most systems that need to run multiple vision understanding tasks on the same video stream. 
{
    \small
    \bibliographystyle{ieeenat_fullname}
    \bibliography{egbib}
}


\end{document}